\documentclass{article}
\usepackage[utf8]{inputenc}
\usepackage[T1]{fontenc}
\usepackage{hyperref}
\usepackage{url}
\usepackage{booktabs}
\usepackage{amsfonts}
\usepackage{amsmath}
\usepackage{nicefrac}
\usepackage{lmodern}
\usepackage{microtype}
\usepackage{graphicx}
\usepackage{xcolor}
\usepackage[numbers,sort&compress]{natbib}
\usepackage{subcaption}
\usepackage{doi}
\usepackage[margin=1in]{geometry}
\usepackage{float}
\usepackage{enumitem}

\title{Adapting TrOCR for Printed Tigrinya Text Recognition: Word-Aware Loss Weighting for Cross-Script Transfer Learning}

\author{
	Yonatan Haile Medhanie\textsuperscript{1} \and
	Yuanhua Ni\textsuperscript{2,*} \\[6pt]
	\textsuperscript{1}College of Software Engineering, Nankai University, Tianjin, China \\
	\textsuperscript{2}College of Artificial Intelligence, Nankai University, Tianjin, China \\
	\textsuperscript{*}Corresponding author \\[4pt]
	\texttt{\{yonatanhaile2026,yhni\}@mail.nankai.edu.cn} \\
}

\date{}

\begin{document}
	
	\maketitle
	
	\begin{abstract}
		
			Transformer-based OCR models have shown strong performance on Latin and CJK scripts, but their application to African syllabic writing systems remains limited. We present the first adaptation of TrOCR for printed Tigrinya using the Ge'ez script. Starting from a pre-trained model, we extend the byte-level BPE tokenizer to cover 230 Ge'ez characters and introduce Word-Aware Loss Weighting to resolve systematic word-boundary failures that arise when applying Latin-centric BPE conventions to a new script.	The unmodified model produces no usable output on Ge'ez text. After adaptation, the TrOCR-Printed variant achieves 0.22\% Character Error Rate and 97.20\% exact match accuracy on a held-out test set of 5,000 synthetic images from the GLOCR dataset. An ablation study confirms that Word-Aware Loss Weighting is the critical component, reducing CER by two orders of magnitude compared to vocabulary extension alone. The full pipeline trains in under three hours on a single 8 GB consumer GPU. All code, model weights, and evaluation scripts are publicly released.
		
	\end{abstract}
	

	\section{Introduction}
	\label{sec:introduction}
	
	Tigrinya is a Semitic language spoken by more than 10 million people in Eritrea and the Tigray region of Ethiopia~\citep{Ethnologue2024Tigrinya}. It is written in the Ge\textquotesingle ez script, an abugida in which each grapheme represents a consonant--vowel combination. The full Tigrinya character inventory exceeds 230 unique symbols, many of which differ only by subtle strokes or diacritical modifications. Despite its large speaker population and official status in Eritrea, Tigrinya remains poorly served by digital language technologies. Commercial OCR platforms provide no dedicated support for the language.
	
	Recent advances in end-to-end Transformer-based OCR, particularly TrOCR~\citep{Li2023}, have demonstrated that pairing a Vision Transformer (ViT) encoder with a language model decoder enables strong text recognition through transfer learning. This architecture has been successfully adapted to non-Latin scripts including Urdu~\citep{Cheema2024}, Tamil~\citep{Murugesh2025}, Manchu~\citep{Chung2025}, and Spanish~\citep{lauar2024spanishtrocrleveragingtransfer}. For the Ge\textquotesingle ez script family, however, no Transformer-based evaluation exists.
	
	Prior Ge\textquotesingle ez script OCR work has relied on CNN-RNN architectures. Belay et al.~\citep{Belay2021} achieved 0.93\% CER on Amharic using an Attention-CTC model, while Hailu et al.~\citep{Hailu2023} reported 2.32\% CER on Tigrinya with a CRNN trained on over one million synthetic samples. Neither study explored Transformer-based approaches.
	
We address this gap through three contributions:
	
	\begin{enumerate}[nosep]
		\item \textbf{First TrOCR adaptation for the Ge\textquotesingle ez script.} We extend the model's tokenizer and embedding layers to cover 230 Tigrinya characters, enabling character-level recognition of the full syllabary.
		
		\item \textbf{Word-Aware Loss Weighting.} We identify and resolve a systematic failure mode caused by BPE space-marker conventions that block learning at word boundaries when new script tokens are added. This technique is applicable to any cross-script BPE adaptation.
		
		\item \textbf{Public benchmark on consumer hardware.} Training completes in under three hours on a single 8~GB GPU, and all resources are publicly released.
	\end{enumerate}

	\section{Related Work}
	\label{sec:related_work}
	
	\subsection{Transformer-Based OCR}
	
	TrOCR~\citep{Li2023} frames text recognition as an image-to-sequence task, combining a ViT encoder with an autoregressive Transformer decoder. Pre-trained on large image and text corpora, TrOCR achieves state-of-the-art results on major printed and handwritten benchmarks for Latin scripts. The architecture's reliance on pre-trained components makes it particularly amenable to transfer learning, as both components arrive with prior knowledge of visual features and language structure.
	
	Alternative architectures include decoder-only approaches such as DTrOCR~\citep{Fujitake2023}, which have shown competitive performance on English and Chinese benchmarks. Hybrid Swin-Transformer encoders have also been explored for non-Latin scripts~\citep{Murugesh2025, Cheema2024}.
	
	\subsection{Ge\textquotesingle ez Script OCR}
	
	Research on Ge\textquotesingle ez script recognition has focused primarily on Amharic. Belay et al.~\citep{Belay2019} proposed a factored CNN that predicts consonant and vowel components separately, exploiting the script's abugida structure. Belay et al.~\citep{Belay2020} subsequently generated a large-scale synthetic dataset of Amharic text-line images using font-based rendering and degradation techniques. The same group later introduced end-to-end sequence models, progressing from LSTM-CTC~\citep{Belay2020} to a blended Attention-CTC architecture achieving 0.93\% CER on the ADOCR synthetic dataset~\citep{Belay2021}. 
	
	For Tigrinya specifically, Hailu et al.~\citep{Hailu2023} designed an end-to-end CRNN trained on over one million synthetic text-line images, reporting 2.32\% CER without post-processing. This represents the most direct prior work, though it uses a pre-Transformer architecture and a substantially larger training set than ours.
	
	The GLOCR dataset~\citep{Gaim2021} provides multiple Tigrinya text-line corpora including a news subset of 230,000 synthetic samples generated from newspaper text. To the best of our knowledge, GLOCR has not previously been used for Transformer-based OCR benchmarking. 
	
	\subsection{Cross-Script Transfer Learning for OCR}
	
	Several recent studies demonstrate successful Transformer OCR adaptation to non-Latin scripts. Murugesh et al.~\citep{Murugesh2025} replace TrOCR's ViT encoder with a Swin Transformer for Tamil handwriting recognition, achieving 5.44\% CER. Cheema et al.~\citep{Cheema2024} combine a Swin backbone with an mBART-50 decoder for bilingual Urdu--English OCR, reaching 1.1\% CER. Chung and Choi~\citep{Chung2025} fine-tune vision-language models on 60,000 synthetic Manchu word images, maintaining 93.1\% word accuracy on real handwritten documents. Lauar and Laurent~\citep{lauar2024spanishtrocrleveragingtransfer} show that fine-tuning the complete English TrOCR on Spanish outperforms systems that replace only the decoder with a language-specific model, suggesting that cross-attention alignment between encoder and decoder carries valuable structural knowledge.
	
	\subsection{Tokenizer Adaptation for New Scripts}
	
	When pre-trained language models encounter scripts absent from their training vocabulary, characters are fragmented into byte-level sequences or mapped to unknown tokens. Pfeiffer et al.~\citep{pfeiffer-etal-2021-unks} demonstrate that extending the vocabulary with script-specific tokens is necessary for meaningful adaptation. Ogueji et al.~\citep{Ogueji2021} show with AfriBERTa that smaller models focused on African languages can outperform larger multilingual models on downstream tasks, supporting the case for targeted adaptation over reliance on general-purpose multilingual coverage.

	\section{Methodology}
	\label{sec:methodology}
	
	\subsection{Dataset}
	\label{subsec:dataset}
	
We use a 20,000-sample subset from the GLOCR news text-lines corpus~\citep{Gaim2021}\footnote{Dataset available at \url{https://github.com/fgaim/GLOCR}}, which contains 230,000 synthetic line images generated from Haddas Ertra newspaper text. Each sample pairs a grayscale text-line image with its ground-truth transcription. From the original 200,000 training samples in the News subset, we selected the first 10,000 for training, 5,000 samples from the 15,000 validation samples for validation, and 5,000 samples from the 15,000 test samples for evaluation. 

Preliminary analysis reveals a compact and uniform dataset with a mean transcription length of 15.8 characters (standard deviation 1.75), a maximum of 21, and a minimum of 4 characters. Images are single-channel grayscale PNGs with standardised height.

Using 20,000 samples rather than the full 230,000 keeps the experiment manageable while still testing how far transfer learning carries the model under data scarcity. Table~\ref{tab:dataset} summarises the dataset structure.
	
	\begin{table}[H]
		\centering
		\caption{Dataset structure and partitioning (subset of GLOCR news corpus).}
		\label{tab:dataset}
		\begin{tabular}{lcccc}
			\toprule
			\textbf{Split} & \textbf{Samples} & \textbf{Mean Len} & \textbf{Min} & \textbf{Max} \\
			\midrule
			Train      & 10,000 & 15.85 & 4 & 21 \\
			Validation &  5,000 & 15.88 & 4 & 18 \\
			Test       &  5,000 & 15.89 & 4 & 18 \\
			\midrule
			Total      & 20,000 & 15.8  & 4 & 21 \\
			\bottomrule
		\end{tabular}
	\end{table}

	\subsection{Model Architecture}
	\label{subsec:model}
	
	We use \texttt{Microsoft/TrOCR-base-handwritten}~\citep{Li2023} as the primary base model, which pairs a BEiT encoder~\citep{Bao2022} with a RoBERTa-initialised decoder~\citep{Liu2019}. The encoder consists of 12 Transformer layers with 768 hidden dimensions, 12 attention heads, and processes input images as $16 \times 16$ non-overlapping patches at $384 \times 384$ resolution. The decoder has 12 layers with 1024 hidden dimensions and 16 attention heads. A learned projection layer bridges the encoder (768-dim) and decoder (1024-dim) during cross-attention. The complete model has approximately 334 million parameters.
	
	\begin{figure}[H]
		\centering
		\includegraphics[width=0.8\textwidth]{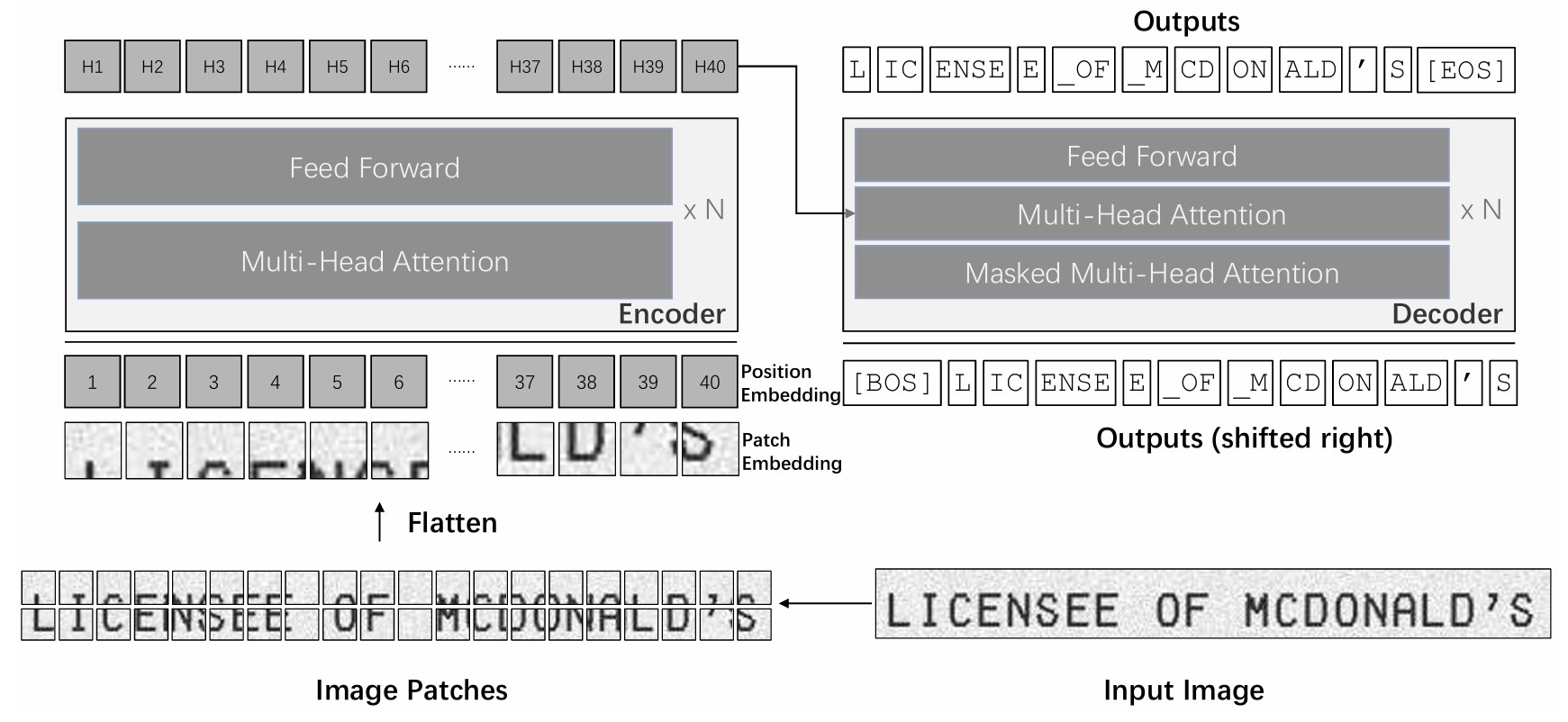}
		\caption{TrOCR architecture overview. The model combines a Vision Transformer encoder for image processing with a Transformer decoder for text generation. (Adapted from~\cite{Li2023})}
		\label{fig:trocr_architecture}
	\end{figure}
	
	The BEiT encoder was pre-trained on ImageNet-21k through masked image modelling, and the full encoder--decoder system was fine-tuned on the IAM Handwriting Database. We selected the handwritten variant as the primary model, reasoning that a model exposed to diverse handwriting styles, irregular spacing, and variable letterforms might adapt more readily to an entirely unfamiliar script.
	
To test this hypothesis, we also fine-tune the printed variant (\texttt{Microsoft/TrOCR-base-printed}) under the same conditions. Both variants share the same architecture and differ only in their Stage~2 fine-tuning data: the handwritten variant was fine-tuned on IAM handwriting, while the printed variant was fine-tuned on synthetic printed text.

	\subsection{Tokenizer Extension}
	\label{subsec:tokenizer}
	
	The original RoBERTa BPE vocabulary of 50,265 tokens was learned predominantly from English text and contains no Ge'ez characters. When applied to Tigrinya text, characters are mapped to \texttt{<unk>} or fragmented into meaningless byte sequences.
	
	We extract all 230 unique characters from the 10,000 training transcriptions, covering the base consonant-vowel combinations (fidel), labialized variants, Ge'ez numerals, and punctuation marks. These are appended to the vocabulary, expanding it from 50,265 to 50,495 tokens. We resize the model's input and output embedding layers accordingly, initialising new embeddings from $\mathcal{N}(0, 0.02)$. We verify encode-decode consistency on 100 randomly selected samples; all reconstruct without information loss.
	
	This approach follows Pfeiffer et al.~\citep{pfeiffer-etal-2021-unks} and treats each Ge'ez character as an atomic token, consistent with the linguistic structure of the script where each fidel represents an indivisible consonant-vowel unit.

	\subsection{Word-Aware Loss Weighting}
	\label{subsec:word_aware_loss}
	
	Even after vocabulary expansion, early experiments revealed a systematic failure, the model consistently dropped the first character after whitespace. The cause is a mismatch between BPE conventions and the new tokens.
	
	RoBERTa's byte-level BPE tokenizer prepends a space marker to word-initial tokens in English (for example, \texttt{\_Word}). The newly added Ge\textquotesingle ez characters enter the vocabulary as isolated tokens with no space-prefixed variants. Because no BPE merge rules have been learned over Ge\textquotesingle ez text, the tokenizer cannot combine a space byte with a Tigrinya character. This creates a blind spot at word boundaries where the decoder fails to learn the transition from space tokens to Ge\textquotesingle ez characters.
	
	\begin{figure}[H]
		\centering
		\includegraphics[width=1\textwidth]{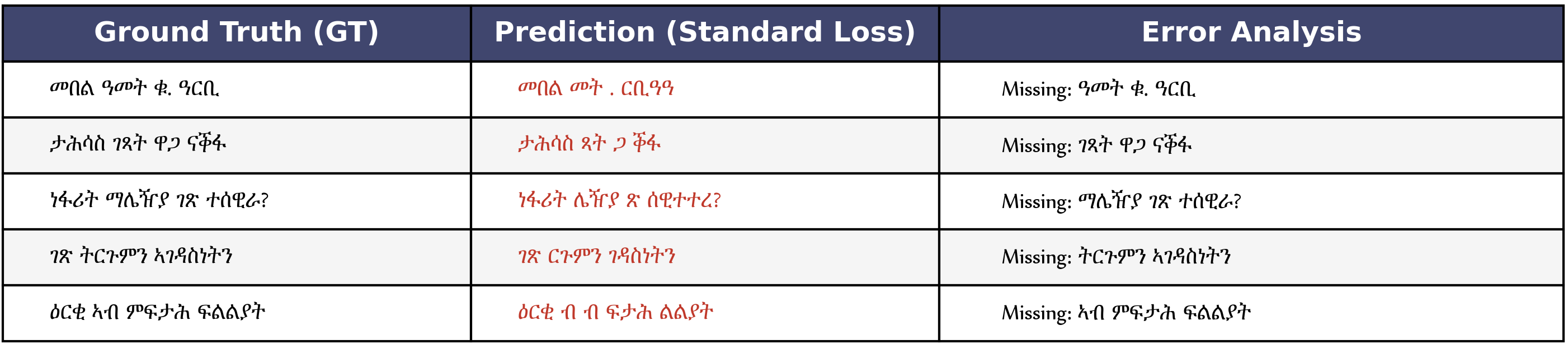}
		\caption{The failure mode observed with standard cross-entropy loss. The model consistently drops the first character of words following a whitespace, reflecting the boundary conflict between the pre-trained BPE tokenizer and the newly added Ge\textquotesingle ez vocabulary.}
		\label{fig:tokenmismatch}
	\end{figure}
	
	To resolve this without retraining the tokenizer from scratch, we introduce \textbf{Word-Aware Loss Weighting}. During initialisation, we scan the vocabulary to identify all tokens containing the BPE space delimiter. During the forward pass, these boundary tokens receive a weight of 2.0 while all other tokens retain a weight of 1.0:
	
	\begin{equation}
		\mathcal{L} = \sum_{i=1}^{N} w_i \cdot \text{CE}(y_i, \hat{y}_i)
		\label{eq:loss}
	\end{equation}
	
	where $\text{CE}$ is the standard cross-entropy between ground-truth token $y_i$ and predicted token $\hat{y}_i$, and $w_i \in \{1.0, 2.0\}$ is the position-dependent weight. Doubling the penalty for errors at word onsets forces the optimiser to learn the space-to-character transition more reliably.

	\subsection{Training Configuration}
	\label{subsec:training}
	
	All model parameters are updated during fine-tuning. The same training procedure is applied identically to both the handwritten and printed variants to enable a direct comparison. We use AdamW~\citep{Loshchilov2019} with a learning rate of $4 \times 10^{-5}$, linear decay to zero, and no warmup. Training runs for 10 epochs with a physical batch size of 2 and gradient accumulation over 4 steps, yielding an effective batch size of 8. Mixed-precision training (FP16) is enabled. Checkpoints are saved every 2,000 steps. The random seed is fixed at 42.
	
	The full training run completes in approximately 2 hours and 40 minutes per variant on a single NVIDIA GeForce RTX 5060 Laptop GPU with 8~GB VRAM. Table~\ref{tab:hyperparams} summarises the configuration.
	
	\begin{table}[H]
		\centering
		\caption{Training hyperparameters.}
		\label{tab:hyperparams}
		\begin{tabular}{ll}
			\toprule
			\textbf{Hyperparameter} & \textbf{Value} \\
			\midrule
			Optimiser & AdamW \\
			Learning rate & $4 \times 10^{-5}$ \\
			LR scheduler & Linear decay (no warmup) \\
			Training epochs & 10 \\
			Per-device batch size & 2 \\
			Gradient accumulation steps & 4 \\
			Effective batch size & 8 \\
			Mixed precision & FP16 \\
			Boundary loss weight & 2.0 \\
			Checkpoint frequency & Every 2,000 steps \\
			Random seed & 42 \\
			Training duration & $\sim$2h 40m (per variant) \\
			\bottomrule
		\end{tabular}
	\end{table}

	\subsection{Evaluation Protocol}
	\label{subsec:evaluation}
	
	Performance is measured using three complementary metrics:
	
	\begin{itemize}[nosep]
		\item \textbf{Character Error Rate (CER):} Based on Levenshtein edit distance~\citep{Levenshtein1966}: $\text{CER} = (S + D + I) / N$, where $S$, $D$, $I$ are character-level substitutions, deletions, and insertions, and $N$ is the reference length.
		\item \textbf{Word Error Rate (WER):} The same edit-distance formula applied at the word level.
		\item \textbf{Exact Match Accuracy:} The proportion of samples with identical predicted and reference transcriptions.
	\end{itemize}
	
	Text is generated using beam search with 5 beams and a maximum sequence length of 128 tokens. Statistical reliability is assessed via bootstrap resampling~\citep{Efron1993} with 1,000 iterations on the test set.

	\section{Results}
	\label{sec:results}
	
	\subsection{Baseline Comparison}
	
	The unmodified TrOCR models fail completely on Tigrinya text. Both the handwritten and printed variants were evaluated on 500 randomly selected test samples in their zero-shot configuration. The handwritten variant produces a CER exceeding 130\% (due to insertion-dominated errors), while the printed variant reaches 99.13\% CER. Neither achieves any exact matches. After fine-tuning with vocabulary extension and Word-Aware Loss Weighting, both variants shift from total failure to functional recognition (Table~\ref{tab:baseline}).
		
	On the 500-sample subset, the printed variant performs slightly better than the handwritten one, with 0.16\% CER versus 0.24\%. On the full 5,000-sample test set, the printed variant reaches 0.22\% CER and 97.20\% accuracy, while the handwritten variant reaches 0.38\% CER and 96.86\% accuracy. For comparison, we also trained a CRNN-CTC model from scratch on the same data, which reaches 0.12\% CER. The lower error rate is useful as a benchmark, but the main result here is that TrOCR can be adapted successfully to Ge\textquotesingle ez script with only tokenizer extension and Word-Aware Loss Weighting. What follows focuses on the TrOCR adaptation in detail.
	
	\begin{table}[H]
		\centering
		\caption{Baseline comparison ($n = 500$ subset and $n = 5{,}000$ full test set).}
		\label{tab:baseline}
		\begin{tabular}{llccc}
			\toprule
			\textbf{Model} & \textbf{Eval Set} & \textbf{CER (\%)} & \textbf{WER (\%)} & \textbf{Acc.\ (\%)} \\
			\midrule
			Handwritten (zero-shot) & $n=500$ & 130.01 & 112.40 & 0.00 \\
			Printed (zero-shot) & $n=500$ & 99.13 & 102.18 & 0.00 \\
			\midrule
			Handwritten (fine-tuned) & $n=500$ & 0.24 & 0.84 & 97.00 \\
			Printed (fine-tuned) & $n=500$ & 0.16 & 0.61 & 97.80 \\
			\midrule
			Handwritten (fine-tuned) & $n=5{,}000$ & 0.38 & 1.15 & 96.86 \\
			Printed (fine-tuned) & $n=5{,}000$ & 0.22 & 0.87 & 97.20 \\
			\midrule
			CRNN-CTC baseline & $n=5{,}000$ & 0.12 & 0.57 & 98.20 \\
			\bottomrule
		\end{tabular}
	\end{table}
	
		This pattern is consistent with recent evidence that TrOCR's visual and structural representations transfer effectively across scripts. Lauar and Laurent~\citep{lauar2024spanishtrocrleveragingtransfer} showed that fine-tuning the complete English TrOCR on Spanish outperforms decoder-only replacement, suggesting that cross-attention alignment carries transferable structural knowledge. Strobel et al.~\citep{Strobel2022} demonstrated that TrOCR adapts to non-English Latin-script historical manuscripts with minimal training data. Our results extend this finding to a fundamentally different writing system, where the script shares no characters with the pre-training language.
	
	\begin{figure}[H]
		\centering
		\includegraphics[width=1\textwidth]{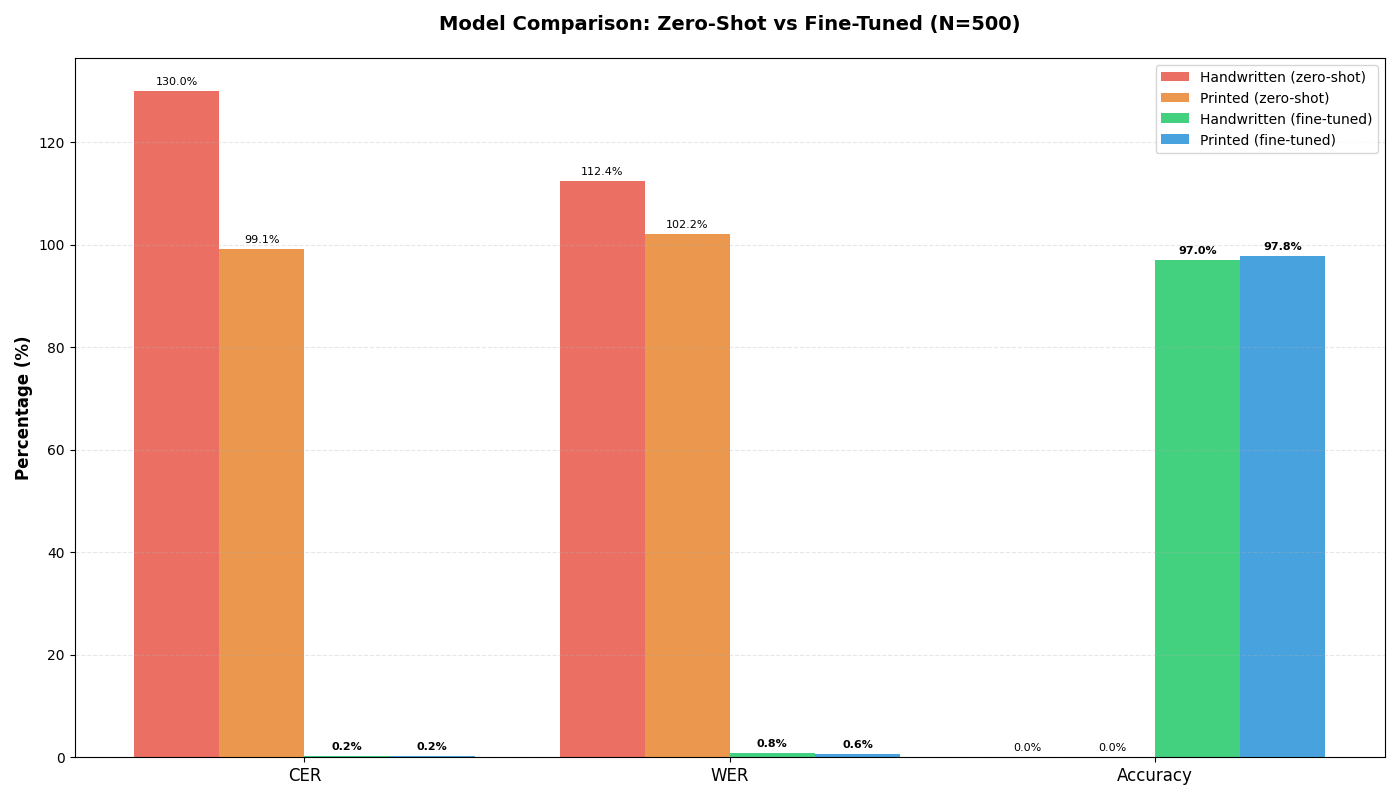}
		\caption{Zero-shot failure versus fine-tuned performance, showing the effect of vocabulary extension and Word-Aware Loss Weighting.}
		\label{fig:baseline_comparison}
	\end{figure}

	\subsection{Effect of Word-Aware Loss Weighting}
	
	To isolate the contribution of the weighted loss, we compare standard fine-tuning (vocabulary extension only) against fine-tuning with Word-Aware Loss Weighting on the full 5,000-sample test set. The ablation uses the handwritten variant to demonstrate the effect of Word-Aware Loss Weighting (Table~\ref{tab:ablation}).
	
	\begin{table}[H]
		\centering
		\caption{Ablation: effect of Word-Aware Loss Weighting ($n = 5{,}000$).}
		\label{tab:ablation}
		\begin{tabular}{lccc}
			\toprule
			\textbf{Approach} & \textbf{CER (\%)} & \textbf{WER (\%)} & \textbf{Accuracy (\%)} \\
			\midrule
			Standard training     & 20.06 & 79.03 & 0.02 \\
			Word-Aware Weighting  &  0.38 &  1.15 & 96.86 \\
			\midrule
			\textbf{Improvement}  & \textbf{$-$98.1\%} & \textbf{$-$98.5\%} & \textbf{+96.84 pp} \\
			\bottomrule
		\end{tabular}
	\end{table}
	
	Without the weighted loss, vocabulary extension alone yields 20.06\% CER and near-zero accuracy. The model systematically omits the first character after whitespace, and this misalignment propagates through subsequent tokens. With Word-Aware Loss Weighting, CER drops from 20.06\% to 0.38\% and accuracy rises to 96.86\%. This demonstrates that the BPE boundary mismatch, rather than vocabulary extension alone, is the critical bottleneck. This two-order-of-magnitude improvement represents the core technical contribution of this work and is directly applicable to any BPE-based cross-script adaptation scenario.

	\subsection{Full Test Set Performance}
	
The handwritten variant reaches 0.38\% CER on the held-out test set, or roughly one character error every 263 characters. The printed variant reaches 0.22\% CER and 97.20\% accuracy. Given the average line length of 15.8 characters, most lines are transcribed without error by all three models. WER is 1.15\% for handwritten TrOCR, 0.87\% for printed TrOCR, and 0.57\% for the CRNN-CTC baseline. Inference latency is about 0.20 seconds per line on the RTX 5060 Laptop GPU.
	
Training was stable. The loss dropped from 33.38 to 0.0013, while validation loss fell from about 0.26 at step 2,000 to around 0.02 at step 12,000. There were no spikes or oscillations, which suggests that the weighted loss behaved as intended and that the model generalized reasonably well.
	
	\begin{figure}[H]
		\centering
		\includegraphics[width=1\textwidth]{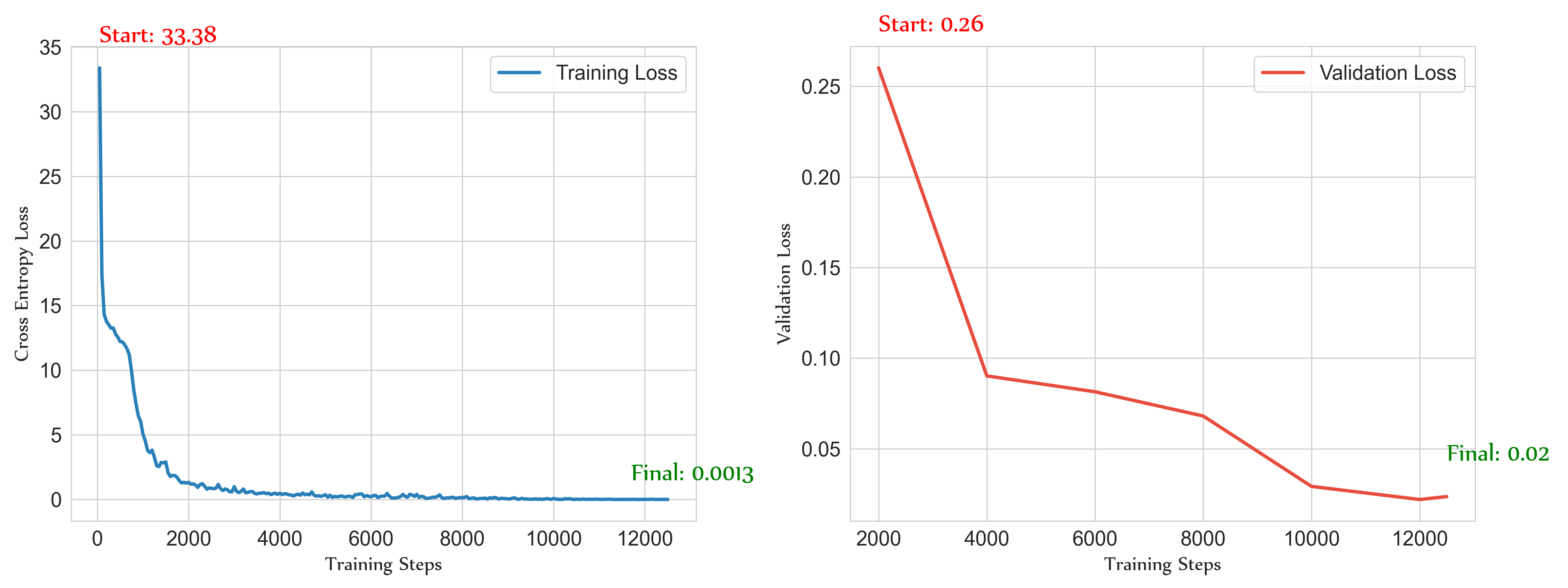}
		\caption{Training and validation loss curves during fine-tuning, showing smooth convergence of both losses and effective generalization.}
		\label{fig:loss_curves}
	\end{figure}

	\subsection{Bootstrap Confidence Intervals}
	
	Bootstrap resampling with 1,000 iterations confirms that the performance estimates are stable under sampling variation on this test set (Table~\ref{tab:bootstrap}).
	
	\begin{table}[H]
		\centering
		\caption{Bootstrap 95\% confidence intervals for the TrOCR-Printed variant (best validation checkpoint).}
		\label{tab:bootstrap}
		\begin{tabular}{lcc}
			\toprule
			\textbf{Metric} & \textbf{Point Estimate} & \textbf{95\% CI} \\
			\midrule
			CER       & 0.20\% & [0.17\%, 0.24\%] \\
			WER       & 0.77\% & [0.64\%, 0.91\%] \\
			Accuracy  & 97.42\% & [96.96\%, 97.84\%] \\
			\bottomrule
		\end{tabular}
	\end{table}
	
	\begin{figure}[H]
		\centering
		\includegraphics[width=1\textwidth]{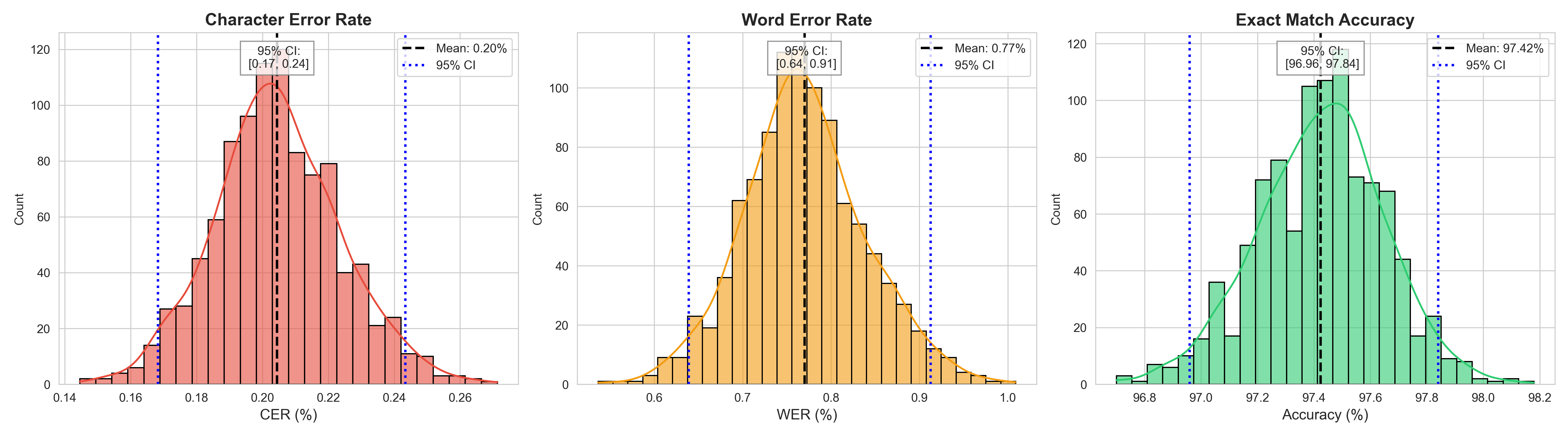}
		\caption{Bootstrap 95\% confidence intervals for the best checkpoint of the TrOCR-Printed variant ($n = 5{,}000$, 1,000 iterations).}
		\label{fig:bootstrap}
	\end{figure}
	
	The CER confidence interval spans only 0.07 percentage points, indicating that the point estimates for the TrOCR-Printed variant are precise for this specific evaluation corpus. The bootstrap mean (0.20\% CER) differs slightly from the single-pass test evaluation (0.22\% CER) due to sampling variance across the 1,000 bootstrap iterations. These intervals characterize measurement precision rather than generalization to unseen domains, fonts, or document conditions; evaluating robustness to such variation is left to future work.

	\subsection{Error Analysis}
 Of 5,000 test samples evaluated on the printed variant, 4,860 (97.20\%) are transcribed perfectly. The remaining 140 error cases were classified using an automatic analyzer that applies linguistic rules based on the Ge'ez script structure. The classifier groups errors by comparing ground-truth and predicted strings according to six categories: characters within the same consonant family differing only in vowel order (diacritic confusions), consonants with labialized variants, visually similar characters from different families, digit and punctuation recognition, spacing and boundary transitions, and mixed-script sequences. Each error is assigned to the most specific applicable category based on edit distance and Unicode character properties (Table~\ref{tab:errors}).
	\begin{table}[H]
		\centering
		\caption{Error distribution across the test set ($n = 5{,}000$).}
		\label{tab:errors}
		\begin{tabular}{lcc}
			\toprule
			\textbf{Error Category} & \textbf{Frequency} & \textbf{Distribution (\%)} \\
			\midrule
			Digits and mixed-script errors & 54 & 1.08 \\
			Visual character substitutions & 28 & 0.56 \\
			Diacritic confusions & 25 & 0.50 \\
			Labialized character errors & 14 & 0.28 \\
			Punctuation errors & 12 & 0.24 \\
			Boundary and spacing errors & 7 & 0.14 \\
			\midrule
			\textbf{Total errors} & \textbf{140} & \textbf{2.80} \\
			\bottomrule
		\end{tabular}
	\end{table}
	
	
The largest source of error is digits and mixed-script text, which accounts for 54 cases. In these samples, the model often confuses Latin digits with visually similar Ge\textquotesingle ez characters or produces the wrong digit sequence. That likely reflects the low frequency of numeric tokens in the training data. Diacritic confusions mostly involve vowel-order mistakes within the same consonant family, where the visual differences are only small strokes that can be blurred by rendering variation. Visual character substitutions occur between distinct consonant families with nearly identical shapes. Boundary and spacing errors are rare, which suggests that Word-Aware Loss Weighting mitigated the boundary failure mode.

\begin{figure}[H]
		\centering
		\begin{subfigure}[b]{0.33\textwidth}
			\centering
			\includegraphics[width=\textwidth]{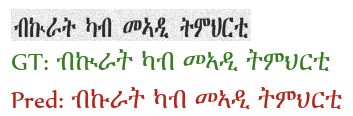}
			\caption{Labialized character errors}
			\label{fig:example_Labialized}
		\end{subfigure}
		\hfill
		\begin{subfigure}[b]{0.32\textwidth}
			\centering
			\includegraphics[width=\textwidth]{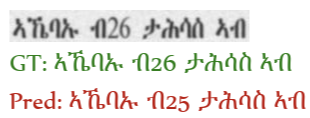}
			\caption{Digits and mixed-script errors}
			\label{fig:example_numeral}
		\end{subfigure}
		\hfill
		\begin{subfigure}[b]{0.32\textwidth}
			\centering
			\includegraphics[width=\textwidth]{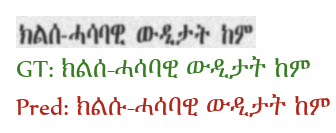}
			\caption{Diacritic confusions}
			\label{fig:example_diacritic}
		\end{subfigure}
		
		\caption{Sample test set error predictions. (a) A labialized character recognition error. (b) Digits and mixed-script errors (c) A diacritic confusion between vowel orders of the same consonant family.}
		\label{fig:sample_predictions}
	\end{figure}

	Representative erroneous predictions from the test set are shown in Figure~\ref{fig:sample_predictions}. These examples illustrate the three most frequent error types observed.

	\section{Discussion}
	\label{sec:discussion}
	
	\textbf{Visual features transfer robustly across scripts.} A ViT encoder pre-trained on natural images and fine-tuned on English handwriting transferred well once the decoder could represent Ge\textquotesingle ez characters. The vision backbone did not need any special redesign; the model learned the printed Tigrinya lines once the tokenizer and loss were fixed. That pattern is consistent with earlier OCR transfer-learning work, including studies on Urdu, Tamil, Manchu, and Spanish~\citep{Cheema2024, Chung2025, Murugesh2025, lauar2024spanishtrocrleveragingtransfer}.
	
	\textbf{Tokenizer conventions are a hidden bottleneck.} The main bottleneck lies at the tokenizer and decoder interface. The model could read the characters, but the decoder's space-marking convention still disrupted word starts. Vocabulary extension alone was not enough; the loss had to place extra weight on boundary tokens before the model stopped dropping the first character after whitespace.
	
	\textbf{Pre-training domain has a measurable but modest impact after adaptation.} Fine-tuning both variants under identical conditions reveals a consistent advantage for the printed variant (0.22\% CER versus 0.38\% CER for the handwritten variant), which suggests that the pre-training domain has a measurable though modest effect on final performance. At the zero-shot level, both variants fail completely (0\% accuracy). Neither pre-training domain helps without tokenizer adaptation. Once the vocabulary is extended and Word-Aware Loss Weighting is applied, the printed variant's closer alignment with the target domain gives it a small but consistent advantage. A CRNN-CTC baseline trained on the same data achieves 0.12\% CER and 98.20\% accuracy, outperforming both TrOCR variants on this synthetic corpus and providing a useful reference point for future architectural comparisons. However, this comparison is based on single training runs; repeating the experiment across multiple random seeds would strengthen the conclusion.
	
	\textbf{Architectural comparison and contribution scope.} While a CRNN-CTC model trained from scratch on the same data achieves lower error rates (0.12\% CER), the primary contribution of this work lies in demonstrating successful cross-script transfer learning with TrOCR and in introducing Word-Aware Loss Weighting as a general solution to BPE boundary mismatches when adapting to new scripts. The architectural advantages of Transformer-based OCR, particularly on degraded documents, handwritten text, and low-resource scenarios, have been documented extensively for Latin scripts~\citep{Li2023}. Our work establishes that these benefits can extend to Ge'ez through appropriate tokenizer adaptation, providing both a viable approach and a methodological template for other non-Latin scripts.
	
	\textbf{The technique generalises beyond Tigrinya.} The same boundary issue may arise whenever a byte-level BPE tokenizer is adapted to a script whose word boundaries do not align with the original tokenization scheme. The weighted-loss fix is simple enough to merit testing on other scripts, although that broader claim still needs direct evidence.
	
	\textbf{Comparison with prior work.} To our knowledge, our best result (0.22\% CER on synthetic printed text) represents the first Transformer-based evaluation on Ge\textquotesingle ez script. Hailu et al.~\citep{Hailu2023} reported 2.32\% CER with a CRNN on over one million Tigrinya samples, though direct comparison is not meaningful due to different datasets, rendering procedures, and evaluation protocols. Our contribution is methodological: we demonstrate that TrOCR's pre-trained visual and linguistic representations transfer to a fundamentally different writing system when the tokenizer is properly adapted.
	
	\textbf{Limitations.} Several limitations remain. The evaluation is synthetic and printed, so we still do not know how the models behave on real scans or handwritten text. We also do not compare against other architectures on exactly the same data, do not vary the boundary weight, and do not run multiple random seeds. The error analysis is informative, but it depends on automatic Unicode-based grouping, so a few mistakes may be assigned to the wrong category.

	\section{Conclusion}
	\label{sec:conclusion}
	
We present the first reported adaptation of TrOCR to the Ge'ez script. The printed variant achieves 0.22\% CER and 97.20\% exact match accuracy on held-out synthetic Tigrinya text. The main contribution of this work is the successful cross-script transfer methodology and the introduction of Word-Aware Loss Weighting, which addresses BPE tokenizer boundary mismatches and reduces CER by two orders of magnitude compared to vocabulary extension alone.
	
 Training finished in under three hours on a single consumer GPU with 8~GB VRAM, which keeps the computational requirements modest for a small research group. All training code, evaluation protocols, and fine-tuned model weights are publicly available.\footnote{Code: \url{https://github.com/YoHa2024NKU/Tigrinya_TrOCR_Printed}} The handwritten variant (\url{https://huggingface.co/Yonatanhaile2026/tigrinya-trocrhandwritten}) and the printed variant (\url{https://huggingface.co/Yonatanhaile2026/tigrinya-trocrprinted}) are hosted on the Hugging Face Model Hub.
	
Future work should test the models on real scans and handwritten Tigrinya, examine whether the same Word-Aware Loss Weighting trick helps in other non-Latin scripts, compare against CNN-RNN baselines on matched data, and check learning curves to determine the minimum training set size required for effective adaptation.

	\section*{Reproducibility}
	
	The GLOCR dataset is publicly available on GitHub~\citep{Gaim2021}. The code repository documents the exact subset selection procedure, data splits, and all hyperparameters including framework default values. Training was conducted with a fixed random seed (42) on the hardware and software stack specified in Table~\ref{tab:environment}.
	
	\begin{table}[H]
		\centering
		\caption{Computational environment.}
		\label{tab:environment}
		\begin{tabular}{ll}
			\toprule
			\textbf{Component} & \textbf{Specification} \\
			\midrule
			GPU & NVIDIA RTX 5060 Laptop (8~GB GDDR7) \\
			CPU & Intel Core i9-14900HX \\
			RAM & 32~GB \\
			OS  & Windows 11 Pro 24H2 \\
			Python & 3.10.11 \\
			PyTorch & 2.6.1 \\
			Transformers & 4.40.0 \\
			CUDA & 12.8 \\
			\bottomrule
		\end{tabular}
	\end{table}

	\section*{APPENDIX}
	\appendix
	
	\section{Ge\textquotesingle ez Script Character Matrix}
	\label{app:fidel}
	
\begin{figure}[H]
		\centering
		\includegraphics[width=0.8\textwidth]{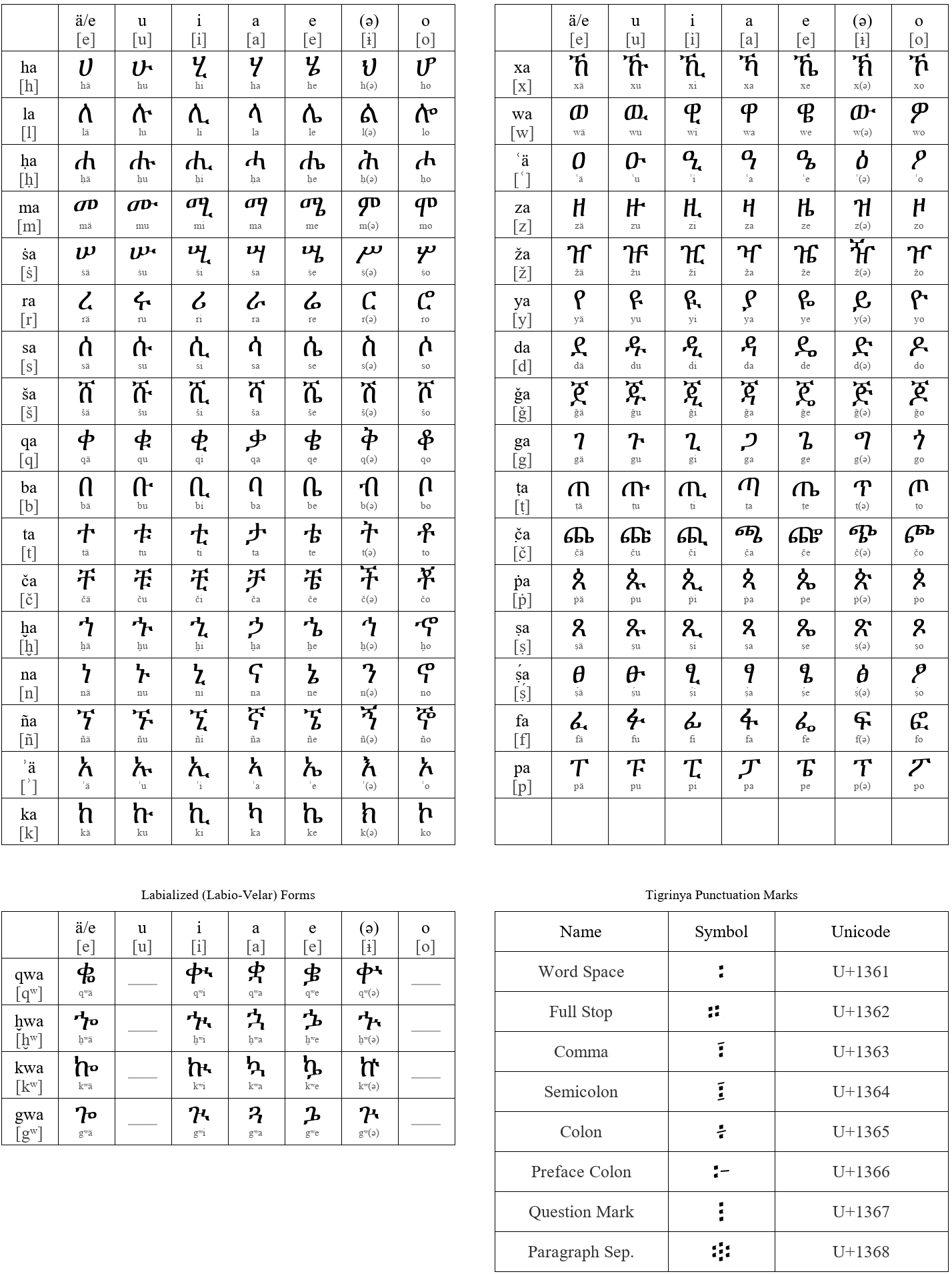}
		\caption{Tigrinya Ge\textquotesingle ez fidel matrix showing 33 base consonants with 7 vowel orders (231 syllographs), 4 labialized consonant groups (20 forms), and 8 punctuation marks.}
		\label{fig:fidel_matrix}
	\end{figure}

	The Ge\textquotesingle ez script is an abugida in which each base consonant has seven vowel-order variants. Figure~\ref{fig:fidel_matrix} shows the core character matrix used in Tigrinya. The visual similarity between vowel orders of the same consonant family and between certain distinct consonant families illustrates the fine-grained discrimination required for accurate recognition.
	
	\bibliographystyle{unsrtnat}
	\bibliography{references}
\end{document}